\documentclass{article}

\usepackage{arxiv}
\usepackage[utf8]{inputenc} 
\usepackage[T1]{fontenc}    
\usepackage{hyperref}       
\usepackage{url}            
\usepackage{booktabs}       
\usepackage{amsfonts}       
\usepackage{nicefrac}       
\usepackage{microtype}      
\usepackage{graphicx}
\usepackage{doi}
\usepackage{subcaption}
\usepackage{tabularx}

\usepackage{bm}  
\usepackage{amsmath}
\usepackage[nameinlink, noabbrev]{cleveref} 
\usepackage{natbib}
\usepackage{scalerel}

\usepackage{multirow}
\usepackage{enumitem}   

\title{Talking like Piping and Instrumentation Diagrams (P\&IDs)}

\author{ 
    \href{https://orcid.org/0000-0001-6382-8196}{\includegraphics[scale=0.06]{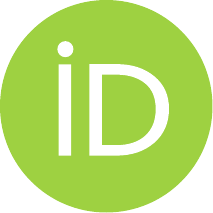}\hspace{1mm}Achmad Anggawirya Alimin} \\
	Process Intelligence Research Group\\
	Department of Chemical Engineering\\
	Delft University of Technology\\
    \And
    \href{https://orcid.org/0000-0002-7837-055X}{\includegraphics[scale=0.06]{figures/orcid.pdf}\hspace{1mm}Dominik P. Goldstein} \\
	Process Intelligence Research Group\\
	Department of Chemical Engineering\\
	Delft University of Technology\\
    \And
    \href{https://orcid.org/0000-0001-7494-9110}{\includegraphics[scale=0.06]{figures/orcid.pdf}\hspace{1mm}Lukas Schulze Balhorn} \\
	Process Intelligence Research Group\\
	Department of Chemical Engineering\\
	Delft University of Technology\\
	\And
    \href{https://orcid.org/0000-0001-8885-6847}{\includegraphics[scale=0.06]{figures/orcid.pdf}\hspace{1mm}Artur M. Schweidtmann}\thanks{corresponding author} \\
	Process Intelligence Research Group\\
	Department of Chemical Engineering\\
	Delft University of Technology\\
    \texttt{A.Schweidtmann@tudelft.nl} \\
}

\renewcommand{\shorttitle}{Talking like P\&IDs}

\hypersetup{
pdftitle={Talking like P\&IDs},
pdfauthor={A. Anggawirya Alimin},
}

\begin{document}
\maketitle

\begin{abstract}
We propose a methodology that allows communication with Piping and Instrumentation Diagrams (P\&IDs) using natural language. In particular, we represent P\&IDs through the DEXPI data model as labeled property graphs and integrate them with Large Language Models (LLMs). The approach consists of three main parts: 1) P\&IDs are cast into a graph representation from the DEXPI format using our pyDEXPI Python package. 2) A tool for generating P\&ID knowledge graphs from pyDEXPI. 3) Integration of the P\&ID knowledge graph to LLMs using graph-based retrieval augmented generation (graph-RAG). This approach allows users to communicate with P\&IDs using natural language. It extends LLM’s ability to retrieve contextual data from P\&IDs and mitigate hallucinations. Leveraging the LLM's large corpus, the model is also able to interpret process information in P\&IDs, which could help engineers in their daily tasks. In the future, this work will also open up opportunities in the context of other generative Artificial Intelligence (genAI) solutions on P\&IDs, and AI-assisted HAZOP studies.
\end{abstract}


\section{Introduction}
\label{sec:introduction}
Piping and Instrumentation Diagrams (P\&IDs) are pivotal in process engineering, serving as comprehensive references across multiple disciplines~\citep{toghraei2019piping}. For process engineers, daily tasks rely on accurate information retrieved from P\&IDs.  For example, engineers require accurate information and interpretation of P\&IDs during design, HAZOP studies, and operation. However, the intricate nature of  P\&IDs and the system's complexity pose challenges to extracting information accurately and efficiently. The current approach mainly relies on manually tracing processing lines on PDF or Computer Aided Engineering (CAE) documents, which is time-consuming and prone to errors, leading to loss of production time and potentially hazardous situations.

Recent digitalization efforts in process engineering have brought a more data-centric approach to P\&ID development. This advancement, often referred to as "smart" or "intelligent" P\&ID, overlays a database on top of process diagrams. Also, recent works proposed the digitization of paper-based P\&IDs into smart P\&IDs~\citep{theisen2023digitization}. While each CAE software may use distinct formats, initiatives to improve interoperability are underway. For instance, the Data Exchange in the Process Industry (DEXPI) initiative established a unified, machine-readable format for P\&IDs~\citep{Theissen2021}. Leveraging this data exchange standard, our group has developed the pyDEXPI Python package, which implements the DEXPI information model for Python~\citep{Goldstein2025}. These developments lay essential groundwork; however, a practical, intuitive implementation is still needed to harness this technology fully and enhance the work experience for process engineers.

Large language models (LLMs) have shown exceptional potential in understanding and processing natural language. The release of GPT-4, which was fine-tuned using reinforcement learning from human feedback (RLHF), marked a significant advancement in tasks such as question-answering and general reasoning~\citep{openai2024gpt4technicalreport}. Additionally, retrieval-augmented generation (RAG) systems have made notable progress in reducing hallucinations by enabling LLMs to query information from databases~\citep{zhao2024retrieval}. However, applying these techniques to P\&IDs presents unique challenges and has not been done yet. As an engineering diagram, P\&ID represents the data in both domain and lexical forms. Domain data represents hierarchical information on equipment, piping, and instrumentation, i.e., tubular heat exchangers consisting of a nozzle, shell, and tube bundle. Meanwhile, lexical data captures the interaction relationships between components, i.e., a valve is connected to a tank to regulate inlet flow, while also being connected to an actuator that manipulates the valve opening. This dual nature of P\&IDs, involving both structured hierarchy and interlinked relationships, makes effective information retrieval particularly complex. 

Building on advancements in process engineering digitalization and LLMs, this work focuses on developing a knowledge graph database derived from P\&IDs and its retrieval method. Specifically, DEXPI P\&IDs will be transformed into Neo4j labeled property graphs using pyDEXPI. A graph-based RAG technique will then be employed to extract and present relevant information. To ensure LLMs are aware of essential contextual information within P\&IDs, high-level representations of P\&IDs will be generated and used as context for retrieval. This results in a more intuitive and streamlined approach for process engineers to interact with and retrieve information from P\&IDs. Moreover, the integration of P\&IDs into foundation models is a key enabler technology for future developments of GenAI applications such as auto-correction of P\&D and AI-assisted HAZOPs~\citep{schweidtmann2024generative}.   

\section{Method}
\label{sec:Method}

\begin{figure}
    \centering
    \includegraphics[width=1\linewidth]{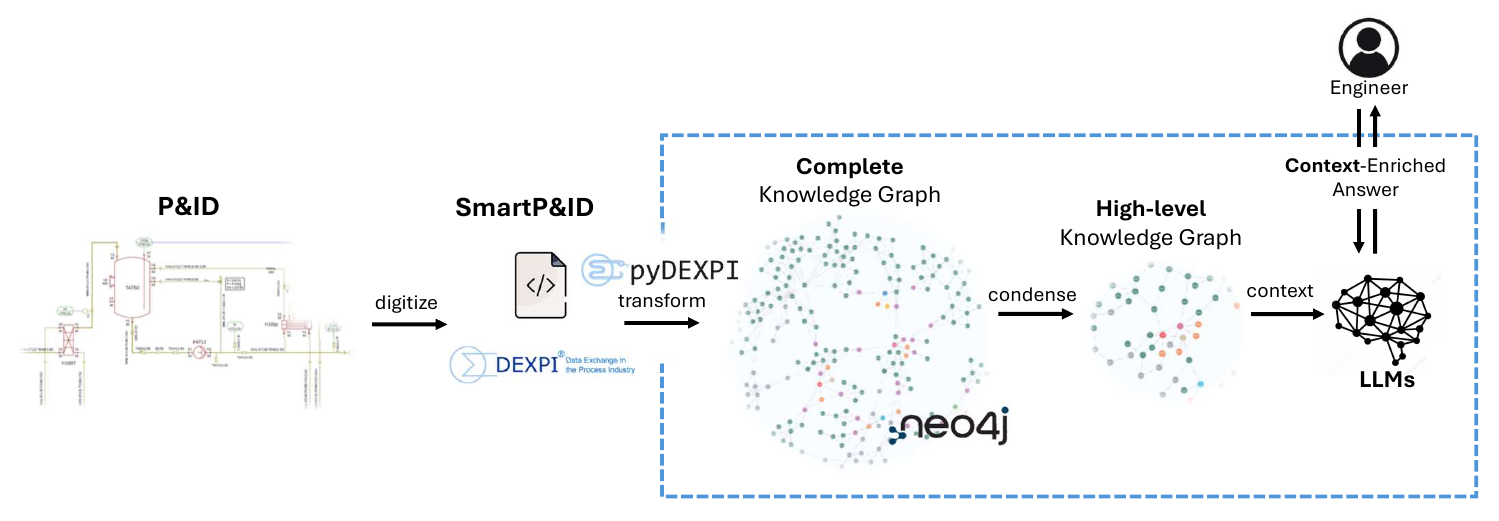}
    \caption{This methodology (inside the dashed box) serves as the final element, developed on top and integrated into digitization and contextualization framework. The process begins by digitizing P\&ID into smart P\&ID using flowsheet digitization tools. The smart P\&ID is then transformed into a knowledge graph using the pyDEXPI package. This graph is further condensed into a high-level knowledge graph, which provides contextual input to an LLM for generating context-enriched responses for interaction with the engineer or user.}
    \label{fig:Framework}
\end{figure}

This section discusses each of the steps in this methodology, including graph representation, knowledge graph generation, and retrieval method as depicted in \Cref{fig:Framework}. A DEXPI P\&ID sample [3] is used as a case study to demonstrate our approach. 

\subsection{P\&ID graph representation}
\label{subsec: PID graph representation}

A graph is an index-free adjacency database consisting of nodes and edges. Nodes represent entities within the dataset, while edges define the relationships between these entities.  Due to its flexible structure and intuitive expression, graph databases are widely used in real-world applications, such as social networking, recommendation engines, fraud detection, and knowledge graphs. Graph databases also demonstrated their effectiveness in process engineering to model entire systems like P\&IDs~\citep{morbach2007ontocape,eibeck_j-park_2019}.

In our data pipeline, smart P\&ID in DEXPI format serves as the input, which is parsed using the pyDEXPI package from its XML file into pyDEXPI class instances. The pyDEXPI class framework implements the DEXPI information model in Python; it can be used as a Python package. The graph representation is created by constructing a NetworkX graph from the class instances as pyDEXPI P\&ID. The pyDEXPI classes modeled as nodes in this graph representation pertain to the following pyDEXPI packages: 1) equipment, 2) piping, and 3) instrumentation. Meanwhile, edges between these elements are derived from two specific relationships of the pyDEXPI P\&ID: domain relationships, which show a child node to the parents, and lexical relationships, which capture the interaction between nodes.  The output is a graph showing the network of connection between class instances.

\subsection{P\&ID knowledge graph generation}
\label{subsec: PID knowledge graph generation}

\begin{figure}
    \centering
    \includegraphics[width=0.9\linewidth]{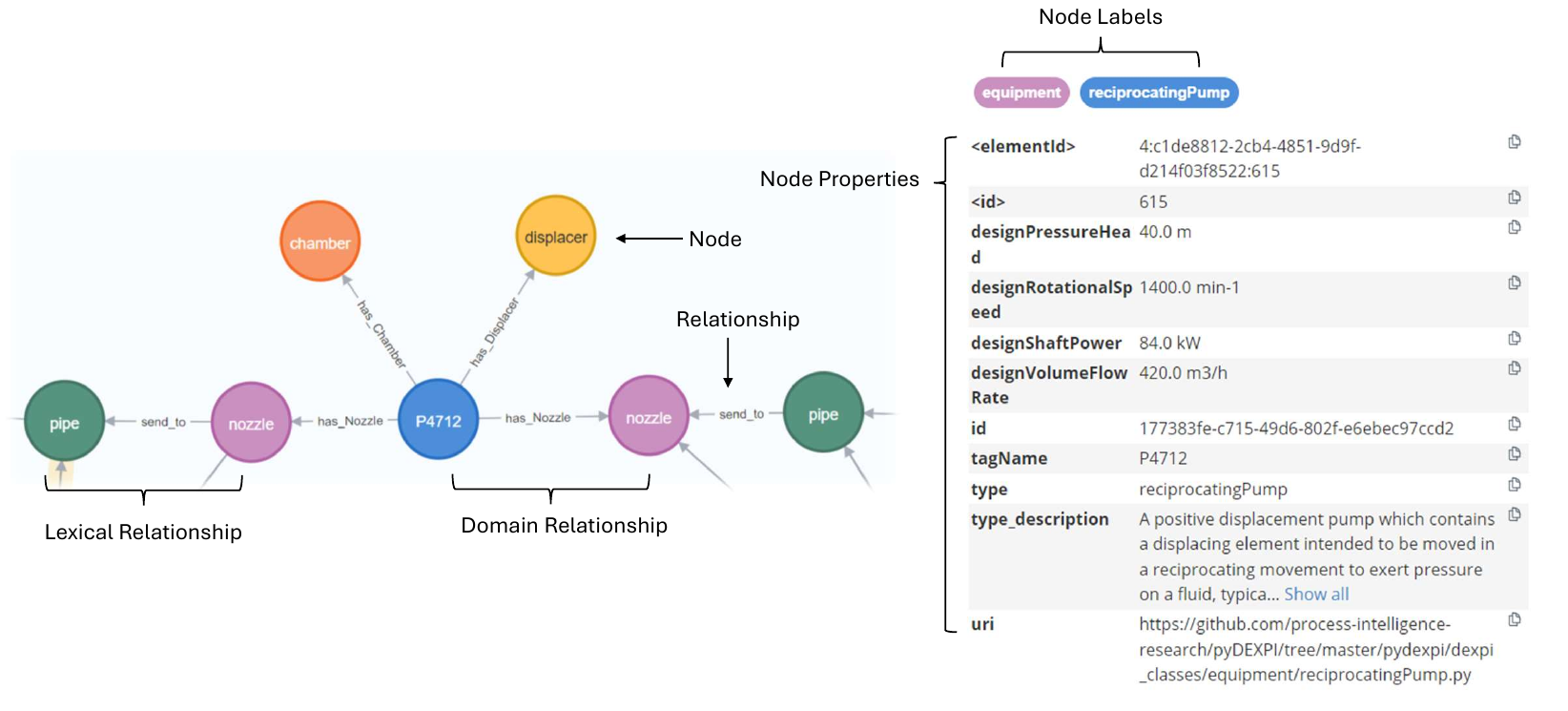}
    \caption{LPG structure for a sample of a pump system P\&ID. (left) Graph pump node with four child nodes: two nozzles, a displacer, and a pump chamber. The stream enters through the right nozzle and exits at the left. (right) Details of node P4712 with labels of \textit{equipment} and \textit{reciprocatingPump}. The node properties contain the pump's specifications.}
    \label{fig:Domain Graph Sample}
\end{figure}

While the graph captures the structure of the flowsheet, it lacks meaningful semantic context in its raw form. As such, the initial graph representation, comprising Python class instances and their connections, remains abstract and unreadable to LLMs.  Semantic information is added to the graph to transform this structure into a readable and functional knowledge graph. This work utilizes a type of knowledge graph called Label Property Graph (LPG) using the Neo4j framework. The LPG graph classifies nodes using the labels and adds key-value pairs as the node properties. This feature is essential to denote different types of components and accommodate the details from the flowsheet.
LPG uses labels as an attribute to distinguish nodes. In our case, multiple labels are employed to classify P\&ID components. Node labels are inherited from the pyDEXPI class name and its parent class, which aligns with DEXPI specification. For example, at a higher level, nodes are assigned labels based on their pyDEXPI package, i.e., equipment, piping, or instrumentation. At a more granular level, additional labels are applied according to each component's specific classification, i.e., reciprocatingPump and centrifugalPump.  Meanwhile, detailed design information, such as size, operating conditions, tag number, and other specifications, is stored as node properties in key-value pairs. \Cref{fig:Domain Graph Sample} illustrates an example of a label property graph for a pump, labeled as \textit{"equipment"} and \textit{"reciprocating pump,"} with its specifications stored as node properties. 

The relationship labels in the knowledge graph are also specified following the relationship in pyDEXPI P\&ID. Domain relationship labels are remarked as \textit{"has\_[componentType]"} or \textit{"is\_located\_in,"} depending on the direction of the relationship. Meanwhile, the lexical relationship is much more specialized. Material flows were labeled with \textit{"send\_to"}; the control structure relationship is represented using words such as  \textit{"control," "send\_signal\_to," "is\_logical\_end\_of,"} and \textit{"measured\_by"}.

\section{Graph-based retrieval method}
\label{subsec:Graph-based retrieval method}

\subsection{High-level graph context generation}
\label{subsection:High-level graph context generation}
While the P\&ID knowledge graph can be directly used as context for LLMs, extracting information becomes increasingly difficult as the graph's context length and complexity grow. For instance, the illustrative DEXPI flowsheet sample which represents a single page of P\&ID with only one tank, two pumps, and two heat exchangers, consists of 212 nodes and 405 relationships. This translates to approximately 67,000 tokens in LLMs, which is inefficient in terms of token usage. The information density in the knowledge graph must be increased to improve efficiency, making the graph more compact and readable for LLMs. The objective is to create high-level graph representations to allow LLMs to focus on the essential structure. This is similar to how engineers read the flowsheet, which develops understanding from a high-level overview of the flowsheet before jumping into details.
Three steps were implemented to create a high-level graph representation:
\begin{enumerate}
    \item Pruning domain information and connection nodes
    \item Condensing low-information nodes
    \item Removing non-process-related node properties
\end{enumerate}

These steps reduce the overall complexity while preserving key relationships and concepts. The high-level graph reduces the knowledge graph size from 212 to 53 nodes,  405 to 57 relationships, and tokens usage from approximately 67,000 to 9,000 tokens. The performance of this high-level graph as context will later be compared to the initial complete knowledge graph.  

\subsection{Chain and retrieval}
\label{subsection:Chain and retrieval}
The LLM agent was created using the Langchain library. In the chain, the LLM is called using the API key or via a local machine. To enhance the interaction, a memory module is added to allow chat history to become context in LLM’s response.  As LLM may take time to give a complete response, the stream method is used to show the generated token directly as it is available. For retrieval, the high-level knowledge graph from Neo4j is exported into a graphml file containing information about nodes, relationships, and properties. This file content is then added as context to the engineered system prompt in the chain. In the LLMs, the string from the graphml file is tokenized together with the chat history and user question and passed to the LLM model to generate a history-aware and contextual response.

\section{Result and discussion}
\label{sec:Result and discussion}
In this section, we evaluate our approach's ability to retrieve information from the flowsheet. The model responses are assessed from a series of questions at various configurations of LLM and knowledge graph. The LLMs are varied to determine pipeline performance across different model scales. Additionally, to test our graph condensing method's ability to represent the flowsheet, the model responses from the completed and high-level graphs are also compared.

\begin{table}
    \centering
    \caption{Performance comparison of complete and high-level knowledge graph as context across different LLMs. Complete LLMs responses from this study are available in \href{https://zenodo.org/records/14899699?token=eyJhbGciOiJIUzUxMiJ9.eyJpZCI6ImU5MGM5ZDhmLWMyNDAtNDY5YS1iZjE0LTBjNGY5MGI3MmI1MSIsImRhdGEiOnt9LCJyYW5kb20iOiI5MDFkM2M1OTY2ODJjOTlhNjllZTkzMGYwODJhZTUwZiJ9.zMxHzz5SUuqHH4fI-QO97se3Y8xGQ3aClIQhKpJyKF2pxNAOCLxi0Mbr4I2tJNttdGj7YFYPzqny6Rx8prLBWg}{\textbf{supporting document}}.}
    \includegraphics[width=1.0\linewidth]{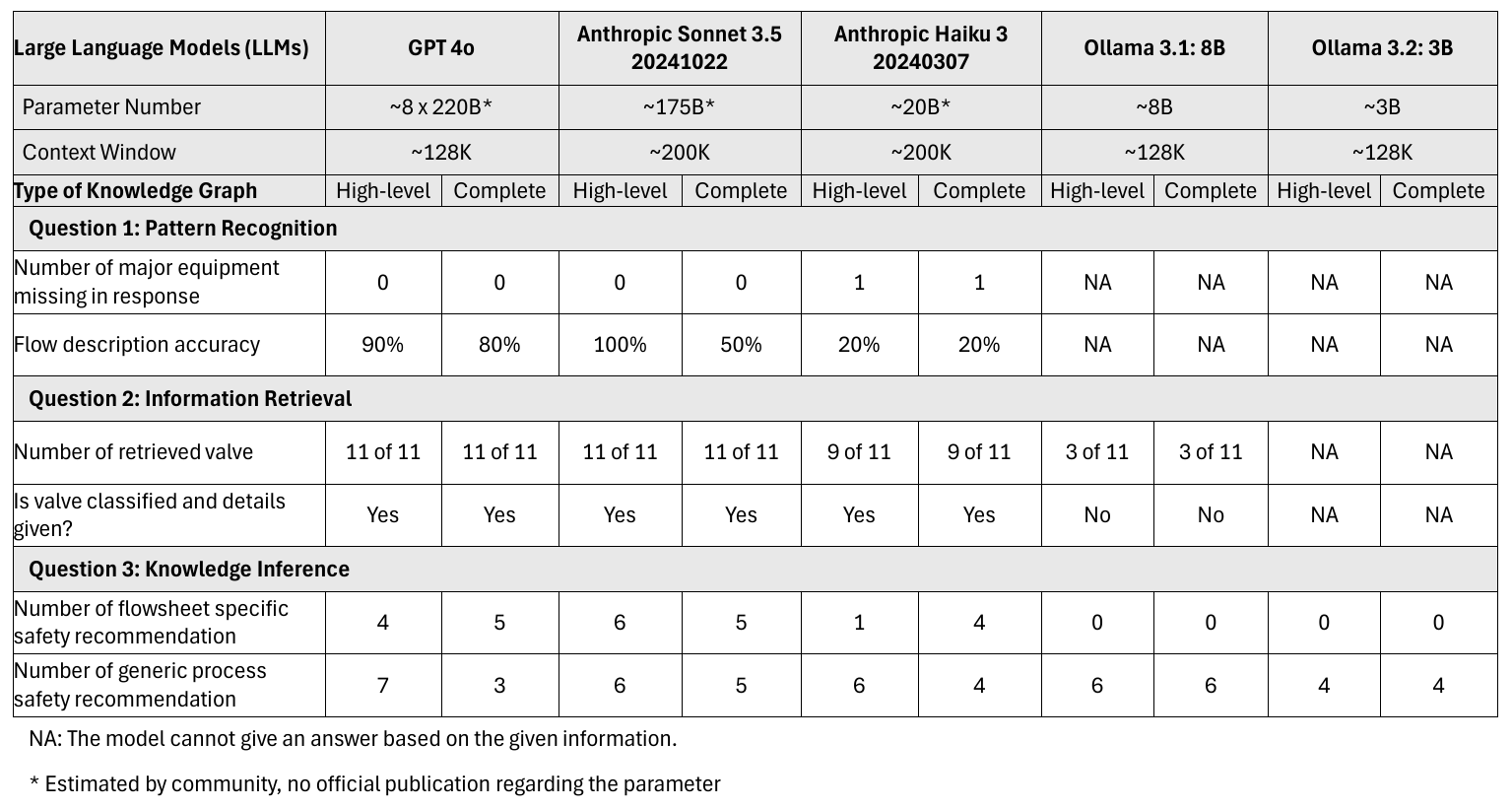}
    \label{table:Table Result}
\end{table}

We designed a series of questions to evaluate distinct aspects, including pattern recognition ability, search completeness, and knowledge inference. The questions are asked sequentially, with each response integrated into the context of the subsequent query to assess how well the model maintains continuity in its analysis. The questions are as follows:
\begin{enumerate}
    \item “Describe the process from inlet to outlet.” This question evaluates the model's pattern recognition ability to extract and infer information from the flowsheet.
    \item "List all valves and their specifications." This question tests the model's capacity for information retrieval and its completeness, specifically its ability to identify and list all valves along with their relevant specifications from the flowsheet.
    \item "Analyze the flowsheet and give recommendations regarding process safety." This question assesses the model's knowledge inference capability, determining whether it can provide insightful process safety recommendations by leveraging its large corpus of information exposed to LLM during training and information from the flowsheet details.
\end{enumerate}

As provided in \Cref{table:Table Result}, the results showed promising potential for this approach to read, search, and infer information from P\&ID. In general, larger LLMs generate more complete and accurate responses. Meanwhile, smaller LLMs have difficulties reading the graph.  Besides the current performance, we also see the potential for smaller LLMs, as they are computationally light and relatively affordable to run on local machines and fine-tune for this specific task.
For pattern recognition, as demonstrated in Question 1, the model’s performance in describing the process is measured by assessing how accurately it follows the flowsheet. A complete score indicates that the model's response correctly traces the process from the inlet to the outlet. A score of 50\% signifies that the model only provided a correct description for the first 50\% of the process, starting from the inlet. This scoring method is chosen because when large language models (LLMs) make an error early in the response, subsequent parts are likely to be incorrect, as each token is generated based on the preceding tokens. On average, using high-level graph representations improved the model’s ability to sequentially describe the process from inlet to outlet by 20\% compared to the complete graph. This infers that our condensing methodology to generate high-level graphs better represents the information to improve LLMs' ability to read the P\&ID.
The response also reflects the performance of large language models (LLMs) which varies significantly depending on the model size. Larger LLMs generally produce more coherent and accurate responses, whereas smaller LLMs are prone to hallucinations or may refuse to answer altogether. A hallucination occurs when a model generates responses that appear plausible but are factually incorrect, nonsensical, or inconsistent with the given prompt. LLMs are probabilistic models that predict the next word in a sequence, and their accuracy depends heavily on the quality and scope of their training data, as well as the model size. LLMs excel at handling widely popular prompts with established consensus but face challenges on controversial topics or topics with limited data~\citep{waldo_gpts_2025}. Additionally, at the cost of computational resources and latency, the model with more parameters tends to have a better performance as it allows for a longer context, which results in better generalization and inferencing~\citep{yang_harnessing_2024}. Since all LLMs used in this study are not fine-tuned to read flowsheets in knowledge graph format, the performance represents more of the size of LLMs. 

\begin{figure}
    \centering
    \includegraphics[width=0.8\linewidth]{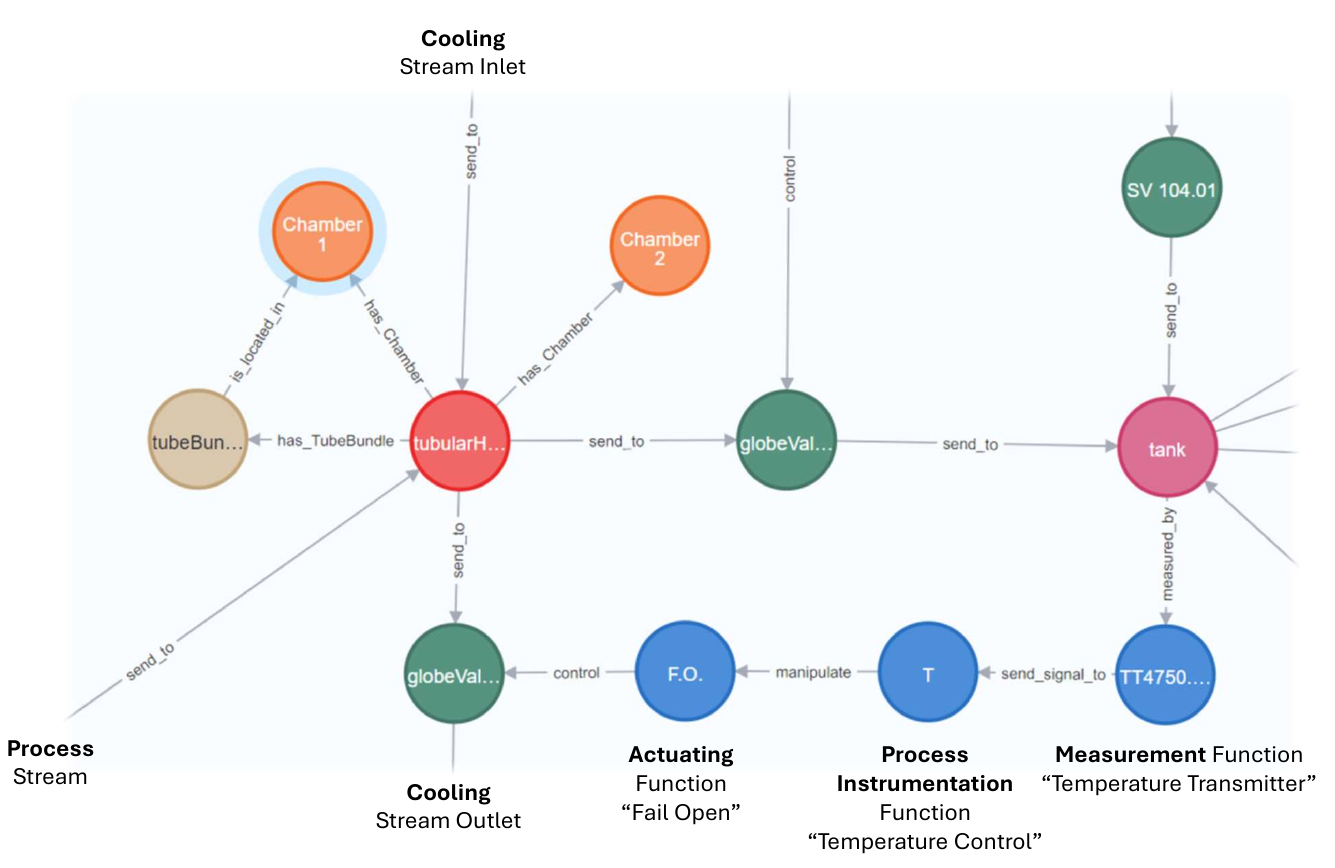}
    \caption{Tank temperature control loop. The tank temperature is measured by the measuring function node TT4750; the measurement signal is then passed to the T process instrumentation function node and actuating function node to the control opening of the globe valve node. The globe valve will control the cooling stream in the tubular heat exchanger to cool the process stream before the tank.}
    \label{fig:Control loop}
\end{figure}

The search completeness results from Question 2, "List all valves and their specifications,"  demonstrated that both types of knowledge graphs produced the equivalent number of valve retrievals. This indicates that the same amount of information from the flowsheet can be effectively represented using the more compact form of a knowledge graph. Additionally, LLMs with a higher number of parameters showed superior retrieval performance. Notably, all 11 valves were retrieved with detailed information, with Anthropic's Sonnet 3.5 model providing particularly better results. On top of the valve, this model also retrieved the control actuators and identified their fail-safe actions. The fail-safe action details were not located directly within the valve node but were instead embedded in the actuating function node, as depicted in \cref{fig:Control loop}. Despite this, the LLM correctly identified and presented this relevant information. Since all necessary details were included in the system prompt as context, the task essentially information recall. 

In the knowledge inference metrics as asked in Question 3, “Analyze the flowsheet and give recommendations regarding process safety,” a model that offers more recommendations mentioning specific components of the flowsheet is highly preferred. This indicates that the LLMs are not only able to retrieve information from the flowsheet but also make the inference to the large corpus on which it has been trained. The evaluation shows that using high-level graphs as contexts generally leads to more flowsheet-specific safety recommendations. This aligns with the improvement in pattern recognition observed in the first question; if LLMs can read the pattern, they also generate more relevant recommendations. Below is a sample of Anthropic’s Sonnet 3.5 process safety recommendations: 

\begin{verbatim}
1. Pressure Safety:
Current Status:
- One spring-loaded safety valve (SV 104.01) with set pressure of 6.0 bar
- Design pressures vary from -1 to 60 bar across equipment
Recommendations:
- Add additional pressure safety valves, especially around:
  * The centrifugal pump P4711 discharge (60 kW pump)
  * The reciprocating pump P4712 discharge (84 kW pump) 
- reciprocating pumps are prone to pressure pulsations
  * Heat exchangers H1007 and H1008 to protect against thermal expansion
- Install pulsation dampeners around the reciprocating pump P4712
- Add pressure indicators and transmitters at more critical points
\end{verbatim}

The results indicate that the Anthropic Sonnet 3.5 is capable of retrieving information on existing safety devices and the design pressure range within the system to formulate relevant recommendations. The recommendations included adding several pressure transmitters at critical locations (though the exact locations were not specified) and installing a pressure safety valve (PSV) on the pump outlet.  While the model suggested a PSV, a rupture disc can also be a choice, depending on the fluid, but the recommendation remains relevant as both devices act as pressure protection devices. In addition, the LLMs also highlighted the importance of addressing pressure pulsation in reciprocating pumps, which is highly pertinent~\citep{Vetter1981}. In the end, the LLMs suggested adding a pressure indicator and transmitter at more critical points. We found this response correct but imprecise; therefore, it’s not particularly useful. Other LLMs like GPT-4o and Anthropic Haiku 3 also give process-specific recommendations in their responses but fewer than Anthropic Sonnet 3.5.

Although the results may appear promising, significant challenges remain to be addressed. The field of process engineering demands low tolerance for mistakes due to the risks associated with processing facilities. At the same time, Gen AI models can exhibit hallucinations and can make errors e.g. missing equipment in the response or imprecise response. Therefore, future work will be focused on improving the accuracy and reliability of the tools. This includes experimenting with retrieval methods, refining the knowledge graph schema, integrating a database, and developing more comprehensive evaluation metrics to assess model performance.

\section{Conclusion}
\label{sec:Conclusion}
We introduced a methodology for intuitive information retrieval and analysis in P\&IDs. The methodology encompasses three steps: 1) graph representation of flowsheets using pyDEXPI, 2) knowledge graph generation, and 3) condensing information to a high-level knowledge graph and retrieval using LLMs. We evaluated the model responses across three different metrics: pattern recognition, search completeness, and knowledge inference. The results demonstrated that this approach enables intuitive information retrieval from P\&ID using LLMs. The LLMs also showed the potential to read, infer, and generate coherent, contextually accurate information from flowsheets. Additionally, the high-level graph condensing method improved the LLM's ability to retrieve the information from P\&ID diagrams. While the current result showed a potential for LLMs to recognize the pattern and infer information from the flowsheet, significant challenges remain to improve the accuracy and reliability in the light of hallucinations of LLMs. 

\section*{Acknowledgement}
\label{sec:Acknowledgement}
We gratefully acknowledge the support provided by Indonesia Endowment Fund for Education Agency LPDP (AAA),  Linde GmbH, Linde Engineering and Siemens Aktiengesellschaft (DPG), as well as Flour© and Top Sector Alliance for Knowledge and Innovation (TKI) by the Dutch Ministry of Economic Affairs and Climate (CHEMIE.PGT.2023.033) (LSB).

\bibliography{references}

\end{document}